%% file: colm2024_conference.tex
\documentclass{article} % For LaTeX2e
\usepackage{colm2024_conference}

\usepackage{microtype}
\usepackage{hyperref}
\usepackage{url}
\usepackage{booktabs}
\definecolor{darkblue}{rgb}{0, 0, 0.5}
\hypersetup{colorlinks=true, citecolor=darkblue, linkcolor=darkblue, urlcolor=darkblue}

\usepackage{adjustbox}
\usepackage{multirow}
\usepackage{booktabs,colortbl,color,xcolor,subcaption}
\usepackage{amssymb}
\usepackage{tabularx}
\usepackage{cleveref}
\usepackage{courier}
\usepackage{wrapfig,lipsum}

\title{\textsc{PairEval}: Open-domain Dialogue Evaluation \\with Pairwise Comparison}

\author{ChaeHun Park \quad Minseok Choi \quad Dohyun Lee \quad Jaegul Choo\\
KAIST AI \\
\texttt{\{ddehun,minseok.choi,aiclaudev,jchoo\}@kaist.ac.kr} \\
}

\colmfinalcopy % Uncomment for camera-ready version, but NOT for submission.
\begin{document}

\maketitle

\begin{abstract}
\input{sections/0_abstract}

\end{abstract}

\input{sections/1_introduction}
\input{sections/2_related_work}
\input{sections/3_method}

\input{sections/4_experiments}

\input{sections/5_conclusion}

\section*{Acknowledgement}
We highly appreciate the reviewers’ insightful comments on our manuscript.

\bibliography{colm2024_conference}
\bibliographystyle{colm2024_conference}

\appendix
\input{sections/A_appendix}

\end{document}

%% file: sections/0_abstract.tex
Building a reliable and automated evaluation metric is a necessary but challenging problem for open-domain dialogue systems. 
Recent studies proposed evaluation metrics that assess generated responses by considering their relevance to previous dialogue histories. 
Although effective, these metrics evaluate individual responses directly rather than considering their relative quality compared to other responses. 
To handle this, we propose  \textsc{PairEval}, a novel dialogue evaluation metric for assessing responses by comparing their quality against responses in different conversations. 
\textsc{PairEval} is built on top of open-sourced and moderate-size language models, and we make them specialized in pairwise comparison between dialogue responses.
Extensive experiments on multiple benchmarks demonstrate that our metric exhibits a higher correlation with human judgments than baseline metrics. We also find that the proposed comparative metric is more robust in detecting common failures from open-domain dialogue systems, including repetition and speaker insensitivity.\footnote{The code and models are available at \url{https://github.com/ddehun/PairEval}.}

%% file: sections/1_introduction.tex
\section{Introduction}
\label{sec:introduction}

Open-domain dialogue systems aim to interact with users by generating natural and engaging responses for a given dialogue history. 
In this field, building an accurate automatic evaluation metric to judge the quality of such generative systems is an important but challenging task \citep{liu-etal-2016-evaluate}.
The difficulties partly originate from the \textit{one-to-many} nature of daily conversations, where a single conversation can be continued with many different follow-up utterances. 
This variety in dialogues makes traditional \textit{reference-based} metrics, which compare generated responses to a limited set of known answers (i.e., references), do not correlate well with human judgments \citep{liu-etal-2016-evaluate}. 
The lack of reliable evaluation metrics impedes real-world deployment of open-domain dialogue systems.

% Open-domain dialogue systems aim to interact with users by generating natural and engaging responses for a given dialogue history. 
% In this field, building an accurate automatic evaluation metric to judge the quality of generations from such systems is an important but challenging task \citep{liu-etal-2016-evaluate}.
% Initial studies had depended on a limited set of known answers (i.e., references) to use \textit{reference-based metrics} such as BLEU \citep{papineni-etal-2002-bleu}, BLEURT \citep{sellam-etal-2020-bleurt}. 
% These methods, however, do not correlate well with human judgments \citep{liu-etal-2016-evaluate}.
% This is due to the \textit{one-to-many} nature of daily conversations, where a single conversation can be continued with many different follow-up utterances.
% Enhanced \textit{reference-based} metrics \citep{liu-etal-2016-evaluate, zhang2019bertscore} which adopt distributed and contextualized representations of neural networks are also limited by the same issues. 

Researchers have proposed various evaluation metrics to achieve reliable assessments of
dialogue systems. Initial studies enhance reference-based metrics by adopting distributed
and contextualized representations by neural networks \citep{liu-etal-2016-evaluate, zhang2019bertscore}.
Subsequent studies introduce \textit{reference-free} metrics that employ prediction models to assess responses based on their relevance to the previous dialogue context~\citep{tao2018ruber,lowe-etal-2017-adem,mehri-eskenazi-2020-usr,zhang-etal-2021-dynaeval,unieval}, showing a higher correlation with human judgments. Moreover, prompting-based metrics have been presented to properly transfer the instruction-following and zero-shot capabilities of large language models~(LLMs) for dialogue evaluation \citep{zhang2023comprehensive,llmeval}.

% To consider \textit{one-to-many} nature of conversation and achieve reliable assessment of dialogue systems, researchers have proposed the \textit{reference-free} metrics. 
% These employ the prediction models to assess responses based on their relevance to the previous dialogue context~\citep{tao2018ruber,lowe-etal-2017-adem,mehri-eskenazi-2020-usr,zhang-etal-2021-dynaeval,unieval} and show a higher correlation with human judgments than \textit{reference-based} metrics.
% However, these works only focused on a ~~

We hold that the dialogue evaluation should aim to assign differentiated scores to responses by considering their relative quality.
In other words, evaluation metrics should make calibrated scores such that they are appropriately aligned with human evaluations. 
Numerous correlation metrics used in the meta-evaluation of evaluation metrics like Spearman~\citep{zar2005spearman} or Kendall rank correlations~\citep{kendall1955rank} reflect this intuition. 
Therefore, we argue that assessing the target responses by considering their relative appropriateness with other ones is a meaningful process to make more reliable evaluation metrics. 
In this regard, several studies evaluate responses by considering their relative quality \citep{sato-etal-2020-evaluating,liusie-comparative}.
However, these approaches usually require exhaustive comparison operations or human-crafted candidate responses, which may not always be available.

In this paper, we propose \textsc{PairEval}, a novel open-domain dialogue evaluation metric with comparative assessments. Our metric assesses the individual responses by comparing their quality against only a limited number of comparison responses. 
Instead of relying on a commercial or proprietary LLM, our metric is built on top of a moderate-size and open-sourced LLM \citep{llama-2-paper}.
To elicit the comparative ability of LMs, we devise a simple but effective learning strategy with a public dialogue corpus.
Experiments on multiple benchmarks show that \textsc{PairEval} outperforms previous evaluation metrics, and sometimes even shows higher performance than metrics with a powerful proprietary LLM. 
Further analysis demonstrates that the pairwise evaluation approach is robust and effective in capturing common failures (e.g., repetitive outcomes) in dialogue systems.

%% file: sections/2_related_work.tex
\section{Related Work}
\label{sec:related_work}
\paragraph{Evaluation Metrics for Open-domain Dialogue Systems}
% Developing accurate and automated metrics is an important but challenging problem in all fields of natural language generation, including open-domain conversation systems. 
Traditional metrics like BLEU~\citep{papineni-etal-2002-bleu} or ROUGE~\citep{lin-2004-rouge} that measure the N-gram overlap between generated responses and answers show a low correlation with human judgments~\citep{liu-etal-2016-evaluate}. 
Several studies use embedding models to consider the semantic similarity between candidate responses and answers~\citep{liu-etal-2016-evaluate,zhang2019bertscore}.
However, these reference-based metrics rely on a set of known answers for similarity comparison, making it hard to consider the wide semantic space of follow-up utterances for a single conversation.  
To tackle this problem, recent studies proposed numerous reference-free metrics that directly predict the relevance of a generated response to the given dialogue history~\citep{tao2018ruber,mehri-eskenazi-2020-usr,mehri2020fed,unieval,full}. 
Specifically, neural classification or regression models are usually trained to distinguish relevant responses from irrelevant ones. 
These predictive metrics have shown meaningful progress along with pre-trained language models~\citep{ghazarian-etal-2019-better, mehri-eskenazi-2020-usr, unieval}, data augmentation strategies~\citep{gupta-etal-2021-synthesizing, park-etal-2021-generating}, and advanced training algorithms~\citep{huang2020grade, zhang-etal-2021-dynaeval}. 
In contrast, our work focuses on a comparative evaluation to consider the relative quality between responses for a more reliable evaluation. 
In open-ended text generation tasks, \citet{pillutla2021mauve} propose a corpus-level evaluation metric, namely \texttt{MAUVE}, which compares model-generated text distributions with human-written ones using divergence frontiers.
To the best of our knowledge, \citet{liusie-comparative} propose a pioneering step that introduces a comparison-based evaluation approach for various natural language generation tasks, including an open-domain dialogue generation task. 
However, they find that exhaustive comparison between candidate responses is needed for reliable quality estimation of dialogues, resulting in an undesirable computational overhead. 
Our work overcomes this challenge by specializing in moderate-size LLMs for pairwise comparison between responses. 
\citet{sato-etal-2020-evaluating} consider the relative rank of generated responses against other false candidates, and a response selection model is employed for this purpose.
Though promising, their evaluation approach requires human evaluation to filter out correct responses from a candidate pool.
In contrast, we show that even a limited number of comparisons are enough to ensure a reasonable correlation with human judgment.

\paragraph{Pairwise Comparison with Large Language Models}
The pairwise comparison approach has been widely explored from various perspectives, including preference learning~\citep{furnkranz2003pairwise}, recommendation~\citep{beutel2019fairness}, reinforcement learning~\citep{xu2020reinforcement}, and retrieval systems~\citep{qin2023large}.
Recent studies that consider the human-aligned behaviors of LMs proposed to reflect human preference in the form of comparisons over multiple model generations \citep{ouyang2022training,bai2022training}.
An LLM-based pairwise evaluation has also been increasingly adopted to build system-level ranking information \citep{boubdir2023elo,zheng2024judging}.
This work evaluates individual responses by comparing their relative quality against the limited number of responses.

%% file: sections/3_method.tex
\input{materials/figure/main_figure}

\section{Method}
\label{sec:method}

We propose \textsc{PairEval}, a new reference-free dialogue evaluation metric based on a pairwise comparison. Our metric assesses responses by comparing their quality against a small number of comparison examples. 
The comparison examples are derived from human-written conversations in a dialogue corpus.
Fig.~\ref{fig:overall} depicts the overall pipeline of \textsc{PairEval}.
% In this section, we first briefly describe the dialogue evaluation task (3.1). We then introduce the overall evaluation pipeline of PairEval (3.2). Finally, we describe the training strategy of LMs to enhance their pairwise comparison ability for more reliable evaluation (3.3).

\subsection{Task Formulation}
\label{sec:method_1}
In this work, we focus on the turn-level and reference-free evaluation of open-domain dialogue systems. 
Given a dialogue history $h$ that consists of multiple utterances between two speakers, a dialogue system outputs a single utterance $r$ as a follow-up response. 
The evaluation metric $M$ considers how suitable the generated response is as the next utterance for the given dialogue history and makes an evaluation score $s=M(h, r)$.
The performance of the metric is usually measured by computing the correlation between human judgments $A=\{a_1, a_2, ..., a_L\}$ and metric scores $S=\{s_1, s_2, ..., s_L\}$ when dialogue systems make $L$ individual responses $R=\{r_1,r_2, ...,r_L\}$ for their dialogue histories $H=\{h_1, h_2,...,h_L\}$.

\subsection{Dialogue Evaluation with Pairwise Comparison}
\label{sec:method_2}
Fig.~\ref{fig:overall} illustrates the evaluation process of \textsc{PairEval}. The metric is designed to judge the quality of a generated response by considering its relative quality against other comparison responses. We use an LLM along with a carefully designed text prompt to perform a pairwise comparison between two responses. Given a pair of conversations in the form of a text prompt, the LM is asked to choose the better response. The probabilities of the target response being better than each comparison example are then aggregated and utilized as its final score. 
The generated response receives higher metric scores as it is predicted as better than the other responses. To construct a group of comparison examples, we use a few randomly sampled conversations from a public dialogue corpus. This makes our metric to be a practical and efficient solution for dialogue evaluation. 

Formally, let $(h_i, r_i)$ denote a target conversation that consists of a generated response $r_i$ under evaluation and its corresponding dialogue history~$h_i$. 
Let $C=\{(h^c_1,r^c_1),(h^c_2,r^c_2),...,(h^c_N,r^c_N)\}$ represent $N$ different comparison examples.
We first make a pair of a target conversation with every comparison example $\{(h^c_j,r^c_j)\}_{j=1}^N$, resulting in $N$ different pairs of conversations. 
Each pair is then converted into an input text of an LLM by replacing the placeholder of our text prompt $x_{ij} = T(h_i, r_i, h^c_j, r^c_j)$. 
Given the text prompt $x_{ij}$, the LM ($\theta$) is asked to choose a response of higher quality. 
To precisely acquire the LM’s prediction in the form of predictive probability, we assign a single label word for each response (i.e., “A” and “B”), and regard the probability of each label word as a score of the response allocated by the LM. 
The probabilities that a generated response $r_i$ is in better quality $s_{ij}=P("r_i$ \textit{is better than} $r_j"|x_{ij}, \theta)$ are stored during every comparison against $c_j$. 
The averaged probability of $r_i$ after every competition is then used as the final evaluation score of the target response. 
For the aggregation of probabilities with different comparison examples, we simply use an average operation~($s_i = \sum_{j=1}^N s_{ij}$). 
In practice, since LLMs are known to be sensitive to ordering in the prompt~\citep{LLM_unfair_eval}, we infer the LMs twice for a single conversation pair with swapped orders and use the averaged probability.

\subsection{Making LMs Specialized in Pairwise Comparison}
\label{sec:method_3}
\input{materials/figure/train_sample}
Although recent LLMs have shown impressive instruction-following and task generalization abilities, applying moderate-size LLMs directly for pairwise dialogue evaluation tasks described in Section~\ref{sec:method_2} may make it hard to ensure reasonable evaluation results. 
For instance, the training examples that require LMs to compare two text inputs and choose the better one would occupy only a small portion of the entire training dataset. 
To tackle this problem, we propose an intuitive training strategy to make LMs specialize in pairwise comparisons for evaluation. 

Specifically, we construct synthetic training examples that instruct an LM to compare two responses of different quality -- \textit{positive} and \textit{negative} responses -- and choose the better one. To obtain a positive response along with its dialogue history, we simply use usual conversations that consist of a dialogue history and its follow-up utterances presented in the ordinary dialogue corpus. 
% To obtain a negative response along with its dialogue history, a dialogue history sampled from the dialogue corpus is used. 
For reliable construction of negative ones, we explore two different types of negative responses with varied difficulties for evaluation: \textit{random} and \textit{adversarial}. The random negative responses are the utterances of different dialogue histories that are arbitrarily sampled from the dialogue corpus. The Adversarial negative responses are written by human annotators to exhibit high superficial relevance to a dialogue context but are not suitable as a follow-up utterance of the context. This adversarial response enables an LM to identify the inappropriateness of a response by capturing more subtle errors rather than relying on the superficial similarity against a dialogue history. In this work, we use human-written negative responses released by \citet{sai-etal-2020-deb} for the adversarial responses. Fig.~\ref{fig:train_sample} illustrates an example of responses with different types.
Given a paired example with positive and negative responses, we construct an input text of an LM with pairwise evaluation prompt $T$.
We then finetune the LM to predict the label word of a positive response correctly.
The location of positive and negative examples in an input text is randomly decided to avoid unintended positional bias.

%% file: materials/figure/main_figure.tex
\begin{figure*}[t]
\centering
\begin{adjustbox}{width=\textwidth}
\begin{tabular}{l}
     \includegraphics[width=0.95\textwidth]{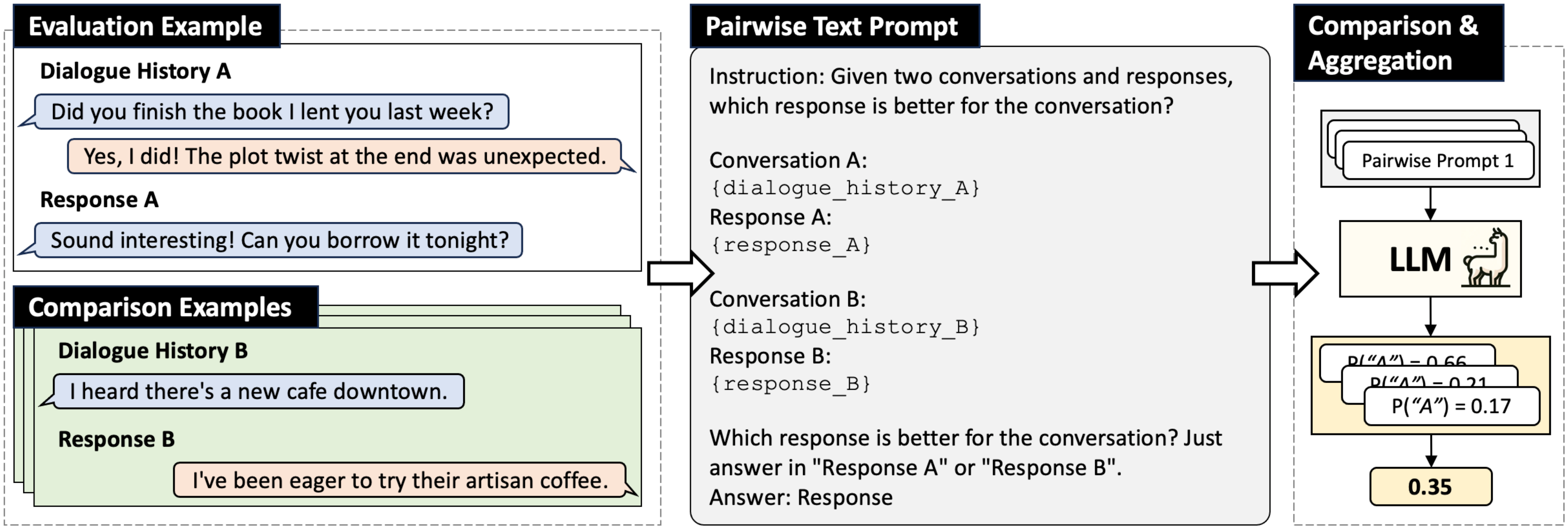}
\end{tabular}
\end{adjustbox}
\caption{The overall illustration of \textsc{PairEval}.}
\label{fig:overall}
% \vspace{-0.5cm}
\end{figure*}

%% file: materials/figure/train_sample.tex
\begin{wrapfigure}{r}{0.4\textwidth}
  \begin{center}
    \includegraphics[width=0.4\textwidth]{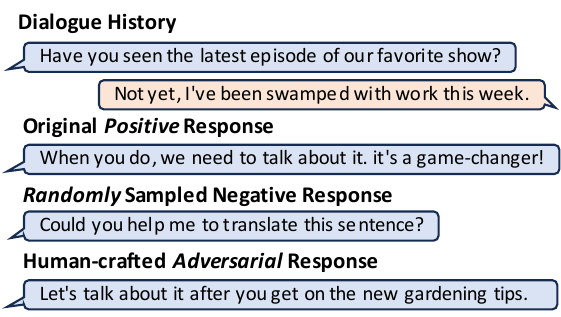}
  \end{center}
  \caption{Responses in different types to finetune a LM in \textsc{PairEval}.
}
\label{fig:train_sample}
\vspace{-0.3cm}
\end{wrapfigure}

%% file: sections/4_experiments.tex
\section{Experiments}
\label{sec:experiments}
The section presents our experimental setups (\cref{secsec:setup}), main results (\cref{secsec:result}), and further analysis~(\cref{secsec:analysis}) as follows.

\subsection{Setup}
\label{secsec:setup}

\subsubsection{Dataset}
\paragraph{Meta-Evaluation Dataset}
We evaluate the performance of each evaluation metric by comparing its outcome scores against scores assigned by human annotators. 
To this end, multiple meta-evaluation datasets with human annotations are used for experiments. Each instance in the dataset consists of (1) dialogue history, (2) generated response from a dialogue system, (3) answer response, and (4) response quality score annotated by human annotators. The details of each meta-evaluation dataset are as follows.

\textit{DailyDialog-Zhao}~\citep{zhao2020designing} consists of 900 evaluation instances from multiple different dialogue systems. The dataset originates from the DailyDialog corpus \citep{li-etal-2017-dailydialog}, serving both as a training corpus of dialogue systems and the source of dialogue histories. The “overall” aspect of the example is used as a human annotation score.
The inter-annotator agreement is measured by Kripendorff's alpha, with a result of 0.8. 
\textit{ConvAI2-USR} and \textit{TopicalChat-USR} are datasets released by \citet{mehri-eskenazi-2020-usr} and consist of 300 and 360 instances, respectively. 
Each dataset is derived by ConvAI2~\citep{dinan2020second} and TopicalChat~\citep{gopalakrishnan2019topical} dialogue corpus, respectively. Three dialogue systems are used as a dialogue system, and the “Overall Quality” score is used in our experiments.
The inter-annotator agreements are measured by Spearman correlation with a result of 0.66 and 0.72 for each dataset. 
\textit{DailyDialog-GRADE} and \textit{ConvAI2-GRADE} are released by \citet{huang2020grade}, each based on the DailyDialog and ConvAI2 datasets. DailyDialog-GRADE and ConvAI2-GRADE both contain 300 instances from two and four systems, respectively.
FED~\citep{mehri2020fed} contains 375 instances of conversations between human speakers or human and dialogue systems. The “Overall” aspect of the examples is used for evaluation.
The inter-annotator agreement is measured by Spearman correlation with a result of 0.82.

\paragraph{Training Dataset}
We finetune an LLM with two widely used open-domain dialogue corpus - \textit{DailyDialog}~\citep{li-etal-2017-dailydialog} and \textit{ConvAI2}~\citep{dinan2020second} - individually, resulting in two different versions of \textsc{PairEval}.
Both corpora contain multi-turn conversations between two speakers, and the number of conversations is 13,118 and 17,818, respectively.
To construct positive response sets, we use randomly sampled utterances along with their previous dialogue history in the train split of a dialogue corpus.
For random negative response sets, we use randomly sampled utterances from a dialogue corpus.
During the training of the LM on the DailyDialog corpus, we also use human-written adversarial responses released by \citet{sai-etal-2020-deb} as an \textit{adversarial} negative type.
The number of training examples is set to 80k and 65k for DailyDialog and ConvAI, respectively.
The LM trained on the ConvAI2 dataset is used for the evaluation of \textit{ConvAI2-GRADE} and \textit{ConvAI2-USR}.
The LM trained on the DailyDialog dataset is evaluated on the remaining meta-evaluation datasets.

\subsubsection{Baselines} 
The following baseline metrics are considered in our experiments.
For better readability, we group them into reference-based and reference-free metrics. Further details are in~\ref{secsec:baseline_detail}.

\paragraph{Reference-based Metrics}
\textit{BLEU-2}~\citep{papineni-etal-2002-bleu} measures the n-gram overlap similarity of a generated response against an answer (reference) response.
\textit{BERTScore}~\citep{zhang2019bertscore} uses contextualized embeddings of a pre-trained LM~\citep{liu2019roberta} for similarity comparison.
\textit{BLEURT}~\citep{sellam-etal-2020-bleurt} is pre-trained on synthetic datasets for a better evaluation of machine translation systems.

\paragraph{Reference-free Metrics}
\textit{USR}~\citep{mehri-eskenazi-2020-usr} is trained to distinguish the original follow-up response of a dialogue history from randomly sampled ones.
\textit{DEnsity}~\citep{park-etal-2023-density} calculates the distance of a generated response from the distribution of relevant responses on the feature space of an LM.
\textit{FED}~\citep{mehri2020fed} and \textit{FULL}~\citep{full} both measure the likelihood of predefined follow-up utterances to estimate the quality of a generated response.
\textit{UniEval}~\citep{unieval} adopt intermediate training on multiple datasets for an evaluation of various natural language generation tasks.
\textit{LLMEval}~\citep{llmeval} is a prompt-based metric that leverages a proprietary LLM~(\texttt{claude-v1.3}) for a dialogue evaluation task.

Besides, we also introduce \textsc{DirectEval}, which uses the same training examples and the LLM as \textsc{PairEval}. 
Instead of performing a pairwise comparison, however, this metric directly predicts the appropriateness of a generated response. Specifically, we ask the LM to predict the quality of a generated response (\textit{“Is the above response a good response to the given conversation?”}), following previous studies~\citep{gupta2022instructdial, liang2023holistic}. The probability of a target response (\textit{P(“Yes”)}) is then used as a metric score. This metric is designed to validate the effectiveness of pairwise comparison by controlling the training configuration to be roughly the same as PairEval. 

\subsubsection{Implementation Details}
We use \texttt{Llama-2-7b-chat}~\citep{llama-2-paper} as an LM of both \textsc{PairEval} and \textsc{DirectEval}.
The LM is finetuned with LoRA~\citep{hu2021lora} for 1 epoch, and the $r$ and $\alpha$ of LoRA is set to 16 and 8, respectively.
AdamW~\citep{adamw} optimizer is used for optimization, and the initial learning rate is set to 1e-4.
The batch size is set to 16.
The number of comparison examples~($N$) is set to 3 and is randomly sampled from the validation set of the DailyDialog corpus. Note that these comparison examples are used across all meta-evaluation datasets.
In all experiments, a single 3090 RTX GPU with 24GB of memory is used.
Regarding the limited computational resources, we use 4-bit quantization \citep{dettmers2024qlora} during the evaluation of \textsc{DirectEval} and \textsc{PairEval}.
% PyTorch~\citep{paszke2019pytorch} and Transformers~\citep{wolf-etal-2020-transformers} frameworks are used in the experiments.

\input{materials/table/1_main}

\subsection{Main Results}
\label{secsec:result}

We use the Pearson correlation coefficient~($r$) and Spearman's rank correlation coefficient~($\rho$) to measure the correlation between human judgments and metric scores.
Table~\ref{tab:main_table} shows the experimental results of all metrics. 
Overall, \textsc{PairEval} achieves the highest performance in four and three out of six datasets in the Pearson and Spearman correlations, respectively.
Our metric works well on dialogues from TopicalChat-USR and FED that are not used during the finetuning stages.
Moreover, \textsc{PairEval} sometimes even shows a higher correlation than LLMEval which relies on a powerful proprietary LLM.
Among the other baseline metrics, UniEval shows reasonable performance.
\textsc{DirectEval}, a metric that uses the same training dataset source as \textsc{PairEval}, also shows competitive performance across multiple benchmarks.
However, it generally performs worse than \textsc{PairEval}, especially in the case of the Pearson correlation coefficient.
We believe these results confirm the effectiveness of the pairwise comparison approach for dialogue evaluation.

\subsection{Analysis}
\label{secsec:analysis}
We conduct further analysis to understand \textsc{PairEval} comprehensively as follows.

\input{materials/table/2_finetune_ablation}

\subsubsection{Impacts of Finetuning on Correlation and Stability of PairEval}
To probe the validity of finetuning described in Section \ref{sec:method_3}, we replace the finetuned LM in \textsc{PairEval} with a pretrained one and observe the results.
For experiments, we use a randomly sampled single conversation from a DailyDialog validation split as a comparison example.
We report the mean and standard deviation of multiple runs~(15) with different comparison examples.
As shown in Table~\ref{tab:finetune_ablation}, finetuning on pairwise evaluation examples greatly contributes to the performance of \textsc{PairEval}.
Furthermore, \textsc{PairEval} with the finetuned LM is more robust to the changes of comparison examples, confirming its validity.

\input{materials/table/3_sample_efficiency}

\subsubsection{Changed Number and Types of Comparison Examples}
We next analyze the impacts of different configurations for comparison examples in \textsc{PairEval} by (1) reducing the number of comparison examples~($N$) and (2) using a few examples in the target meta-evaluation dataset~($Test$) instead of a dialogue corpus~($Random$). We report the averaged results of five runs with different comparison examples. Finetuned LMs are used for the experiments, and results are shown in Table~\ref{tab:comparion_ablation}. We first observe that the increased number of comparison examples usually contributes to a better correlation with human judgments.
Regarding the different types of comparison examples, using the examples in the meta-evaluation datasets usually offers better performance.
However, access to such test examples may not always be available.
Therefore, we believe that using randomly sampled conversations from a dialogue corpus can be a reasonable option.

\input{materials/table/3b_exhaustive}

We also explore the more exhaustive evaluation case, where a single evaluation sample is compared against all other samples in a meta-evaluation dataset to induce the final score.
In other words, if we have $M$ different responses under evaluation, the number of an LM inference becomes $\mathcal{O}(M^2)$. 
Results in Table~\ref{tab:exhaustive} show that conducting an exhaustive comparison between all evaluation examples contributes to the increased correlation. For instance, In the ConvAI2-USR dataset, the correlations of \textsc{PairEval} are increased from 66.5 and 71.1 to 70.0 and 71.9 in Pearson and Spearman correlation, respectively.
Nevertheless, when considering the efficiency of an evaluation process, relying on a few~($N\le3$) comparison examples can be a reasonable choice.

\input{materials/table/4_position_bias}
\subsubsection{Position Bias}
Recent studies have reported that an LLM-based comparative evaluation usually suffers from positional bias, where an LLM exhibits spurious preference over instances in a certain position in the input texts~\citep{LLM_unfair_eval, zheng2024judging}.
To analyze the impacts of such bias on performance, we locate the target evaluation conversation in the $First$ and $Second$ positions of an input prompt respectively, and observe the changed performance.
Besides, we also report the original scores of \textsc{PairEval} ($Both$) that use an averaged probability when the target response is located in $First$ and $Second$ positions.
A single conversation is used as a comparison example for experiments.
As shown in Table~\ref{tab:position_bias}, both zero-shot and finetuned LMs usually suffer from position bias to some extent.
This problem is more significant in a zero-shot setting, where the max value of a correlation difference between the $First$ and $Second$ is 18.1~(41.4 and 23.3 in FED).
The finetuned LM looks relatively free from such issues, where a maximum correlation difference is 4.9 (57.7 and 62.6 in TopicalChat-USR).

\input{materials/table/5_hard_negative}
\subsubsection{Impacts of Human-Written Adversarial Negatives for LM Finetuning}
We verify the effectiveness of human-written adversarial examples described in Section~\ref{sec:method_3}.
Table~\ref{tab:hard_negative} presents the results of \textsc{PairEval} when the LM is finetuned with and without human-written hard negative responses.
For a fair comparison, we use the same number of training examples for both cases by replacing the hard negatives with random ones.
From the results, we confirm that adversarial examples generally contribute to increasing the correlation of \textsc{PairEval} with human evaluation.
One remaining issue is that the creation process of adversarial responses involves a human annotation process, which is not always available and is not scalable.
In this regard, a dataset generation strategy with LLMs can be an efficient and effective alternative~\citep{yoo-etal-2021-gpt3mix-leveraging,liu-etal-2022-wanli,lee-etal-2022-pneg}.
We leave such explorations as our future work.

\input{materials/table/6_robustness}
\subsubsection{Robustness to Adversarially Manipulated Responses}

\citet{khalid-lee-2022-explaining} report that automated metrics for dialogue evaluation often struggle to give higher scores to correct responses rather than to adversarially curated and incorrect ones.
We probe whether our pairwise comparative approach can alleviate such issues and can accurately identify subtly corrupted responses.
To this end, we use a meta-evaluation dataset proposed by \citet{khalid-lee-2022-explaining}.
The dataset is created by manipulating the original human-written response into various corruption strategies.
Each evaluation example in the dataset consists of a (1) dialogue history, (2) original response, and (3) corrupted response.
The metric should assign a higher score to the original response than the corrupted one.
For experiments, we compare the performance of \textsc{PairEval} and \textsc{DirectEval}.
\textsc{PairEval} compares two responses -- original and corrupted -- by locating them in a single input text, while \textsc{DirectEval} evaluates two responses individually. As shown in Table~\ref{tab:robustness}, \textsc{PairEval} achieves higher accuracy in most attack types, confirming its robustness on dialogue evaluation.

\input{materials/figure/scatter}
\input{materials/table/8_case_study}

\subsubsection{Qualitative Results}
\paragraph{Metric Visualization with Scattered Predictions}
Fig.~\ref{fig:scatter} presents a prediction of selected metrics along with human judgments in the DailyDialog-GRADE dataset. 
The x and y axes of each point represent human judgments and metric scores, respectively.
Among the baseline metrics, reference-based metrics like BLEU and BLEURT fail to assign discriminated scores as human ones.
Reference-free metrics like USR and DirectEval show a positive correlation with human judgments.
\textsc{PairEval} shows a strong correlation with human evaluations. 

\paragraph{Case Study}
Table~\ref{tab:case_study} presents two selected examples from meta-evaluation datasets and their evaluation results of automatic metrics and human annotators. For both examples, \textsc{PairEval} evaluates generated responses closely to human scores.

%% file: materials/table/1_main.tex
\begin{table*}[t!]
\centering
\begin{adjustbox}{width=1.0\textwidth}
\begin{tabular}{
    l cc c cc c cc c cc c cc c cc c cc
}
\toprule
\multirow{2}{*}{Methods}
 & \multicolumn{2}{c}{\textbf{\begin{tabular}[c]{@{}c@{}}DailyDialog\\GRADE  \end{tabular}}}  
&& \multicolumn{2}{c}{\textbf{\begin{tabular}[c]{@{}c@{}}DailyDialog\\Zhao\end{tabular}}} 
&& \multicolumn{2}{c}{\textbf{\begin{tabular}[c]{@{}c@{}}ConvAI2\\GRADE\end{tabular}}} 
&& \multicolumn{2}{c}{\textbf{\begin{tabular}[c]{@{}c@{}}ConvAI2\\USR\end{tabular}}} 
&& \multicolumn{2}{c}{\textbf{\begin{tabular}[c]{@{}c@{}}TopicalChat\\USR\end{tabular}}} 
&& \multicolumn{2}{c}{\textbf{\begin{tabular}[c]{@{}c@{}}FED\end{tabular}}}
&& \multicolumn{2}{c}{\textbf{\begin{tabular}[c]{@{}c@{}}Avg.\end{tabular}}}\\

\cline{2-3} \cline{5-6} \cline{8-9} \cline{11-12} \cline{14-15} \cline{17-18} \cline{20-21}

& $r$ & $\rho$ && $r$ & $\rho$ && $r$ & $\rho$ && $r$ & $\rho$ && $r$ & $\rho$ && $r$ & $\rho$ && $r$ & $\rho$ \\
\hline
\rowcolor[rgb]{0.882,0.882,0.882}  \multicolumn{21}{l}{\textit{Reference-based Metrics}} \\

BLEU-2~\citep{papineni-etal-2002-bleu}      &14.6*&	10.3*&&	35.5&	20.6&&	5.41*&	9.64*&&	    32.3&	31.5&&	45.9&	46.2&&	-&	-&&	-&	- \\
BLEURT~\citep{sellam-etal-2020-bleurt}      &17.5*&	12.2*&&	34.1&	28.7&&	16.9*&	16.9*&&	    33.3&	29.9&&	44.7&	41.7&&	-&	-&&	-&	- \\
BERTScore~\cite{zhang2019bertscore}   &12.9*&	9.9*&&	36.4&	29.4&&	26.0&	27.7&&	    33.3&	28.3&&	46.9&	46.2&&	-&	-&&	-&	- \\

\hline
\rowcolor[rgb]{0.882,0.882,0.882}  \multicolumn{21}{l}{\textit{Reference-free Metrics}} \\
FED~\citep{mehri2020fed}&	2.6*&	0.1*&&	-9.0*&	-8.5*&&	-6.1*&	-4.6*&&	-2.0*&	-0.7*&&	-7.1*	&-6.9*&&	11.9*	&9.4* &&-1.6&	-1.9 \\
USR~\citep{mehri-eskenazi-2020-usr}	&27.5&	23.8&&	48.8	&51.6&&	40.3&	40.0&&	60.9&	48.1&&	40.7	&32.5&&	11.4&	11.7&&38.2	&34.6\\
CTC~\citep{ctc}&	-15.1*&	-14.7*&&	-5.6*&	-9.2*&&	3.9*&	4.1*&&	47.4&	48.3&&	39.8&	36.3&&	16.1*	&19.5*&&14.4	&14.0  \\
FULL~\citep{full}&	-8.8*	&-10.0*&&	5.1*&	6.0*	&&22.0*&	23.8*&&	7.0*&	8.9*&&	-4.9*&	-7.2*&&	47.1&	\underline{50.6}&&11.3	&12.0 \\
UniEval~\citep{unieval}&	7.6*&	5.7*&&	30.9&	30.1&&	46.6&	47.9&&	\underline{65.1}&	65.9&&	\underline{62.0}&	64.5&&	29.8&	29.9&& 46.9	&47.7 \\
DEnsity~\citep{park-etal-2023-density}&	30.3&	29.5&&	56.8&	57.0&&	48.0&	48.6&&	57.0&	63.0	&&16.3&	24.7&&	24.5&	21.4&&38.8	&40.7  \\
\textsc{DirectEval} &	\underline{38.6}&	\underline{45.1}&&	\underline{62.4}&	\underline{70.0}&&	\underline{55.5}&	56.2&&	58.3&	\textbf{75.0}&&	39.8&	\textbf{73.9}&&	44.4&	50.4&&\underline{49.8}	&\underline{61.8} \\
\textsc{PairEval} (Ours) &	\textbf{47.8}&	\textbf{55.8}&&	\textbf{62.9}&	\textbf{70.7}&&	51.1&	\underline{56.6}&&	\textbf{66.5}&	\underline{71.1}&&	\textbf{70.8}&	\underline{72.2}&&	\underline{51.9}&	\textbf{52.0}&&\textbf{58.5}	&\textbf{63.1} \\
\hline
\rowcolor[rgb]{0.882,0.882,0.882}  \multicolumn{21}{l}{\textit{Proprietary LLM-based Metrics}} \\
LLMEval~\citep{llmeval} & 34.6&	34.9&&	-&	-&&	\textbf{61.3}&	\textbf{61.8}&&	53.3&	51.5&&	49.0&	49.9&&	\textbf{59.7}&	49.9 &&-&	-  \\
\bottomrule
                   
\end{tabular}
\end{adjustbox}
\caption{
The correlations between automatic metrics and humans on meta-evaluation datasets. $r$ and $\rho$ denote Pearson correlation and Spearman's rank correlation coefficient, respectively.
All values with p $>$ 1e-5 are marked with *.
The highest and the second highest scores in each column are marked in \textbf{bold} and \underline{underline}, respectively.
For \textsc{PairEval}, we report the averaged scores over five runs with different comparison examples.
}
\label{tab:main_table}
% \vspace{-0.3cm}
\end{table*}

%% file: materials/table/2_finetune_ablation.tex
\begin{table*}[t!]

\centering
\begin{adjustbox}{width=0.75\textwidth}
\begin{tabular}{
    l cc c cc c cc c cc 
}
\toprule
\multirow{2}{*}{LM}
 & \multicolumn{2}{c}{\textbf{\begin{tabular}[c]{@{}c@{}}DailyDialog\\Zhao  \end{tabular}}}  
&& \multicolumn{2}{c}{\textbf{\begin{tabular}[c]{@{}c@{}}DailyDialog\\GRADE\end{tabular}}} 
&& \multicolumn{2}{c}{\textbf{\begin{tabular}[c]{@{}c@{}}TopicalChat\\USR\end{tabular}}} 
&& \multicolumn{2}{c}{\textbf{\begin{tabular}[c]{@{}c@{}}FED\end{tabular}}} \\
\cline{2-3} \cline{5-6} \cline{8-9} \cline{11-12} 
& $\mu$ & $\sigma$ &&  $\mu$ & $\sigma$ &&  $\mu$ & $\sigma$ &&  $\mu$ & $\sigma$ \\
\hline
Zero-shot&16.2&	5.0&&	31.1&	3.5&&	42.6&	2.8&&	44.6&	3.6 \\
Fine-tuned&55.5&	1.1&&	70.5&	0.4&&	71.9&	0.5&&	51.2&	1.1 \\
\toprule
                   
\end{tabular}
\end{adjustbox}
\caption{
Mean~($\mu$) and standard deviation~($\sigma$) results of fifteen runs with different random comparison examples $(N=1)$.
}
\label{tab:finetune_ablation}
\vspace{-0.5cm}
\end{table*}

%% file: materials/table/3_sample_efficiency.tex
\begin{table*}[t!]
\small
\centering
\begin{adjustbox}{width=0.9\textwidth}

\begin{tabular}{
    c cc c cc c cc c cc c cc c cc
}
\toprule
\multirow{2}{*}{$N$}
 & \multicolumn{2}{c}{\textbf{\begin{tabular}[c]{@{}c@{}}DailyDialog\\GRADE  \end{tabular}}}  
&& \multicolumn{2}{c}{\textbf{\begin{tabular}[c]{@{}c@{}}DailyDialog\\Zhao\end{tabular}}} 
&& \multicolumn{2}{c}{\textbf{\begin{tabular}[c]{@{}c@{}}ConvAI2\\GRADE\end{tabular}}} 
&& \multicolumn{2}{c}{\textbf{\begin{tabular}[c]{@{}c@{}}ConvAI2\\USR\end{tabular}}} 
&& \multicolumn{2}{c}{\textbf{\begin{tabular}[c]{@{}c@{}}TopicalChat\\USR\end{tabular}}} 
&& \multicolumn{2}{c}{\textbf{\begin{tabular}[c]{@{}c@{}}FED\end{tabular}}} \\

\cline{2-3} \cline{5-6} \cline{8-9} \cline{11-12} \cline{14-15} \cline{17-18}

& $r$ & $\rho$ && $r$ & $\rho$ && $r$ & $\rho$ && $r$ & $\rho$ && $r$ & $\rho$ && $r$ & $\rho$ \\
\hline
\rowcolor[rgb]{0.882,0.882,0.882}  \multicolumn{18}{l}{\textit{Comparison Examples from Dialogue Corpus (Random)}} \\
$N=1$   &42.2           &	\underline{56.1}&&	           57.0&	            70.6&&	48.3&	            56.5                         &&	61.3&	            70.9            &&	64.5&	                        72.1&&	46.2&	50.9 \\
$N=2$   &45.5            &	\underline{56.1}&&	           60.8&	\underline{70.7}&&	49.0&	            56.3                        &&	66.3&	            \underline{71.1}&&	69.9&	            \underline{72.3}&&	50.6&	\underline{\textbf{52.0}} \\
$N=3$   &\underline{47.8}           &	55.8&&	\underline{62.9}&	\underline{70.7}&&	\underline{51.1}&	\underline{\textbf{56.6}}&&	\underline{\textbf{66.5}}&	\underline{71.1}&&	\underline{\textbf{70.8}}&	             72.2&&	\underline{\textbf{51.9}}&	\underline{\textbf{52.0}} \\

\rowcolor[rgb]{0.882,0.882,0.882}  \multicolumn{18}{l}{\textit{Comparison Examples from Meta-Evaluation Dataset (Test)}} \\
$N=1$   &50.9&	55.8&&	57.3&	70.3&&	46.7&	55.8&&	51.9&	71.5&&	60.1&	72.1&&	44.5&	49.9 \\
$N=2$   &54.1&	56.4&&	65.3&	\underline{\textbf{71.0}}&&	50.2&	\underline{\textbf{56.6}}&&	63.7&	\underline{\textbf{72.0}}&&	\underline{70.1}&	72.7&&	\underline{51.1}&	51.0 \\
$N=3$   &\underline{\textbf{54.9}}&	\underline{\textbf{56.5}}&&	\underline{\textbf{67.1}}&	\underline{\textbf{71.0}}&&	\underline{\textbf{53.9}}&    56.2&&	\underline{65.9}&	71.7&&	69.9&	\underline{\textbf{72.8}}&&	50.9&	\underline{51.4} \\

\hline
\end{tabular}
\end{adjustbox}
\caption{
Results with changed number~($N$) and type~(\textit{Random} and \textit{Test}) of comparison examples.
The highest score within the same type of comparison example is \underline{underlined}.
The highest score among all configurations is further highlighted in \textbf{bold}.
All results are averaged scores over five runs with different comparison examples.
}
\label{tab:comparion_ablation}
% \vspace{-0.3cm}
\end{table*}

%% file: materials/table/3b_exhaustive.tex
% \begin{wraptable}{r}{7.0cm}
% \small
% \centering
% \begin{tabular}{l cc c cc }
% \toprule
% \multirow{2}{*}{Methods}
%  & \multicolumn{2}{c}{\textbf{\begin{tabular}[c]{@{}c@{}}DailyDialog\\GRADE  \end{tabular}}}  
% && \multicolumn{2}{c}{\textbf{\begin{tabular}[c]{@{}c@{}}ConvAI2\\USR\end{tabular}}} \\
% \cline{2-3} \cline{5-6} 
% & $r$ & $\rho$ && $r$ & $\rho$  \\
% \hline
% Zero-shot&16.6	&17.3	&&30.1&	29.8\\
% Fine-tuned&55.8	&56.4	&&70.0&	71.9\\
% \toprule
% \end{tabular}
% \caption{Results with an exhaustive comparison between meta-evaluation examples.}
% \label{tab:exhaustive}
% \vspace{-0.3cm}
% \end{wraptable}

\begin{table}[t]
\small
\centering
\begin{tabular}{l cc c cc }
\toprule
\multirow{2}{*}{Methods}
 & \multicolumn{2}{c}{\textbf{\begin{tabular}[c]{@{}c@{}}DailyDialog\\GRADE  \end{tabular}}}  
&& \multicolumn{2}{c}{\textbf{\begin{tabular}[c]{@{}c@{}}ConvAI2\\USR\end{tabular}}} \\
\cline{2-3} \cline{5-6} 
& $r$ & $\rho$ && $r$ & $\rho$  \\
\hline
Zero-shot&16.6	&17.3	&&30.1&	29.8\\
Fine-tuned&55.8	&56.4	&&70.0&	71.9\\
\toprule
\end{tabular}
\caption{Results with an exhaustive comparison between meta-evaluation examples.}
\label{tab:exhaustive}

\end{table}

% \begin{table*}[t!]

% \centering
% \begin{adjustbox}{width=0.75\textwidth}
% \begin{tabular}{
%     l cc c cc c cc c cc 
% }
% \toprule
% \multirow{2}{*}{LM}
%  & \multicolumn{2}{c}{\textbf{\begin{tabular}[c]{@{}c@{}}DailyDialog\\Zhao  \end{tabular}}}  
% && \multicolumn{2}{c}{\textbf{\begin{tabular}[c]{@{}c@{}}DailyDialog\\GRADE\end{tabular}}} 
% && \multicolumn{2}{c}{\textbf{\begin{tabular}[c]{@{}c@{}}TopicalChat\\USR\end{tabular}}} 
% && \multicolumn{2}{c}{\textbf{\begin{tabular}[c]{@{}c@{}}FED\end{tabular}}} \\
% \cline{2-3} \cline{5-6} \cline{8-9} \cline{11-12} 
% & $\mu$ & $\sigma$ &&  $\mu$ & $\sigma$ &&  $\mu$ & $\sigma$ &&  $\mu$ & $\sigma$ \\
% \hline
% Zero-shot&16.2&	5.0&&	31.1&	3.5&&	42.6&	2.8&&	44.6&	3.6 \\
% Fine-tuned&55.5&	1.1&&	70.5&	0.4&&	71.9&	0.5&&	51.2&	1.1 \\
% \toprule
                   
% \end{tabular}
% \end{adjustbox}
% \caption{
% Mean~($\mu$) and standard deviation~($\sigma$) results of fifteen runs with different random comparison examples $(N=1)$.
% }
% \label{tab:finetune_ablation}
% \vspace{-0.5cm}
% \end{table*}

%% file: materials/table/4_position_bias.tex
\begin{table}[t!]
\small
\centering
\begin{adjustbox}{width=0.95\textwidth}
\begin{tabular}{
    ll cc c cc c cc c cc
}
\toprule
\multirow{2}{*}{LM}
& \multirow{2}{*}{Position}
& \multicolumn{2}{c}{\textbf{\begin{tabular}[c]{@{}c@{}}DailyDialog\\Zhao  \end{tabular}}}  
&& \multicolumn{2}{c}{\textbf{\begin{tabular}[c]{@{}c@{}}DailyDialog\\GRADE\end{tabular}}} 
&& \multicolumn{2}{c}{\textbf{\begin{tabular}[c]{@{}c@{}}TopicalChat\\USR\end{tabular}}} 
&& \multicolumn{2}{c}{\textbf{\begin{tabular}[c]{@{}c@{}}FED\end{tabular}}} \\
\cline{3-4} \cline{6-7} \cline{9-10} \cline{12-13} 
&& $r$ & $\rho$ && $r$ & $\rho$  && $r$ & $\rho$  && $r$ & $\rho$  \\
\hline
\multirow{3}{*}{Zero-shot}
&First&
7.4&    7.2&&	24.2&	27.1&&	33.0&	35.5&&	41.4&	47.0 \\
&Second&
15.4&	19.4&&	26.5&	30.2&&	35.4&	40.0&&	23.3&	33.5  \\
&Both&
14.3&	15.9&&	31.2&	32.2&&	39.8&	42.2&&	42.3&	44.5  \\
\hline
\multirow{3}{*}{Finetuned}
&First&40.8&	54.3&&	55.8&	69.5&&	57.7&	73.2&&	45.3&	50.6\\
&Second&40.9&	56.8&&	54.8&	70.8&&	62.6&	70.6&&	45.4&	50.1  \\
&Both&42.2&	56.1&&	57.0&	70.6&&	64.5&	72.1&&	46.2&	50.9  \\

\toprule

\end{tabular}
\end{adjustbox}
\caption{
Analysis of position bias results with a single random comparison example~($N=1$). 
The $Position$ denotes a location of evaluation examples in an input prompt.
We report the averaged results of five runs with different comparison examples.
}
\label{tab:position_bias}
% \vspace{-0.3cm}
\end{table}

%% file: materials/table/5_hard_negative.tex
\begin{table}[t!]
\small
\centering
\begin{adjustbox}{width=0.95\textwidth}
\begin{tabular}{
    cc cc c cc c cc c cc
}
\toprule
\multirow{2}{*}{\begin{tabular}[c]{@{}c@{}}Comparison\\Example \end{tabular}}
& \multirow{2}{*}{\begin{tabular}[c]{@{}c@{}}Hard\\Negative \end{tabular}}
& \multicolumn{2}{c}{\textbf{\begin{tabular}[c]{@{}c@{}}DailyDialog\\GRADE  \end{tabular}}}  
&& \multicolumn{2}{c}{\textbf{\begin{tabular}[c]{@{}c@{}}DailyDialog\\Zhao\end{tabular}}} 
&& \multicolumn{2}{c}{\textbf{\begin{tabular}[c]{@{}c@{}}TopicalChat\\USR\end{tabular}}} 
&& \multicolumn{2}{c}{\textbf{\begin{tabular}[c]{@{}c@{}}FED\end{tabular}}} \\
\cline{3-4} \cline{6-7} \cline{9-10} \cline{12-13} 
&& $r$ & $\rho$ && $r$ & $\rho$  && $r$ & $\rho$  && $r$ & $\rho$  \\
\hline
\multirow{2}{*}{Random}
&&36.9&	39.6&&	62.4&	69.2&&	69.0&	69.9&&	45.3&	45.0 \\
& \checkmark &50.8	&56.4&&	66.4&	70.7&&	68.9&	72.4&&	52.4&	51.6\\
\hline
\multirow{2}{*}{Test}
&&37.5&	38.2&&	63.8&	68.7&&	70.9&	72.5&&	43.4&	43.4  \\
&\checkmark&49.0&	55.5&&	64.5&	71.1&&	71.1&	71.8&&	53.3&	51.8\\

\toprule

\end{tabular}
\end{adjustbox}
\caption{
Results with ablating human-written hard negative for LM finetuning ($N=1$). The $Test$ denotes comparison examples from each meta-evaluation dataset.
}
\label{tab:hard_negative}
% \vspace{-0.3cm}
\end{table}

%% file: materials/table/6_robustness.tex
\begin{table}[t!]
% \begin{adjustbox}{width=0.9\textwidth}
\small
\centering

\begin{subtable}{0.98\textwidth}

\begin{tabularx}{0.98\textwidth}{lccccccccc}
\toprule
 & \multicolumn{9}{c}{Adversarial Attack Type} \\ \cmidrule(r{-5.5pt}){2-10}
\multicolumn{1}{c}{Method}& Entailment
&&\begin{tabular}[c]{@{}c@{}} Pronoun\\Usage\end{tabular}
&&\begin{tabular}[c]{@{}c@{}}Named\\Entities\end{tabular}
&&\begin{tabular}[c]{@{}c@{}}Speaker\\Sensitiveness\end{tabular}
&&Contradict  \\ 
\midrule
\textsc{DirectEval} & \underline{93.9}&&    95.7&&   \underline{94.9}&&   79.0&&   95.5 \\
\textsc{PairEval}& 92.3&&    \underline{98.6}&&   \underline{94.9}&&   \underline{95.6}&&   \underline{96.4} \\
\bottomrule
\end{tabularx}
\caption{Evaluation results with different adversarial attack types.}
\end{subtable}

\bigskip 

\begin{subtable}{0.98\textwidth}
\begin{tabularx}{0.98\textwidth}{lcccccc}
\toprule
& \multicolumn{6}{c}{Adversarial Attack Type} \\ 
\cmidrule(r{-10.5pt}){2-7}
\multicolumn{1}{c}{Method}&Repetition
&\begin{tabular}[c]{@{}c@{}}Vocabulary\\Diversity\end{tabular}
&\begin{tabular}[c]{@{}c@{}}Bad\\Paraphrase\end{tabular}
&Entrainment
&Dullness
&	\textbf{Avg.} \\ \midrule

\textsc{DirectEval}      &	71.9&	79.9&	91.9&	82.1&	\underline{97.5}& 88.2 \\
\textsc{PairEval}&  	\underline{98.4}&	\underline{97.0}&	\underline{95.1}&	\underline{96.2}&	97.1& \underline{96.2} \\

\bottomrule
\end{tabularx}
\caption{\textit{(cont'd)} Evaluation results with different adversarial attack types.}
\end{subtable}
\caption{
\textbf{Accuracy of Different Metrics on Adversarial Attack} The accuracy is higher if the original response receives a higher score than a corrupted one by a metric. The higher score for each column is highlighted in \underline{underline}.
}
\label{tab:robustness}
% \vspace{-0.3cm}
% \end{}
\end{table}

%% file: materials/figure/scatter.tex
\begin{figure}[t!]
\centering
\includegraphics[width=0.98\textwidth]{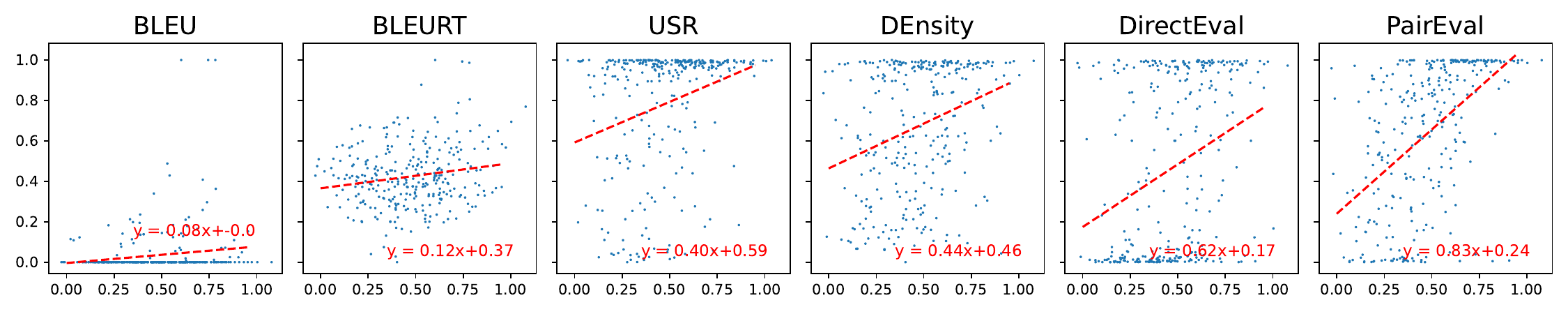}
\caption{
Scatter plots between human judgments and metric scores on the DailyDialog-Grade dataset. 
Each point indicates a response, and the x and y values of each point indicate denote human and metric scores, respectively. 
We add a noise sampled from $\mathcal{N}(0, 0.03^2)$ to human scores for better visualization. 
The red line indicates a linear regression.
}
\label{fig:scatter}
% \vspace{-0.3cm}
\end{figure}

%% file: materials/table/8_case_study.tex
\begin{table}[t!]
\centering
\tiny
\begin{adjustbox}{width=0.95\textwidth}
\begin{tabular}{cccccccc}
\toprule
\textbf{History} & \multicolumn{7}{l}{
\begin{tabular}[l]{@{}l@{}} 
% A: I would love to have that incentive to write and publish a book.\\
% B: Well you should do a children's book then, because they will buy 1,500 of them \\
% \quad instead of just the 1,000 lol \\
A: Children's books require less writing and more illustrations. It would be worth it for the \\
\quad extra 500 copies though\\
B: Haha I agree with you. Do you ever buy stuff from Amazon? do you know why Jeff Bezos\\
\quad put a clock into a mountain lol\\
\end{tabular}
} \\ \hline \\[-5pt]

\textbf{Reference} & \multicolumn{7}{l}{
\begin{tabular}[l]{@{}l@{}} 
A: I buy things from Amazon all the time. Isn't that the clock that's supposed to run for 10,000 years 
%\quad 
\end{tabular}
} \\ \hline  \\[-5pt]

\textbf{Response} & \multicolumn{7}{l}{
A: I have not, what is it about?
} \\ \hline \\[-5pt]

\textbf{Evaluation} & Human & BLEU-2 & BERTScore & FULL & UniEval & DirectEval & PairEval \\[1pt]
Score(Rank) & 3.0(0.56)&  0.0(0.95)&  0.84(0.82)&0.76(0.72)  &0.76(0.72)  &0.99(0.78)  &0.47(0.65)  \\
 \bottomrule
 %% \\[-5pt]
 \hline \\[-5pt]
 \textbf{History} & \multicolumn{7}{l}{
\begin{tabular}[l]{@{}l@{}} 
A: What do you think about the equipment in our company?\\[1pt]
B: We have to equip our company with a network of work stations at every office and show room\\[0.2pt]
\quad as soon as possible. \\
\end{tabular}
} \\ \hline \\[-5pt]

\textbf{Reference} & \multicolumn{7}{l}{
\begin{tabular}[l]{@{}l@{}} 
A: I also realized this issue. And office automation is essential.
\end{tabular}
} \\ \hline \\[-5pt]

\textbf{Response} & \multicolumn{7}{l}{
A: How do you ensure quality?
} \\\hline \\[-5pt]

\textbf{Evaluation} & Human & BLEU-2 & BERTScore & FULL & UniEval & DirectEval & PairEval \\[1pt]
Score(Rank) & 3.9(0.09)&0.0(1.0)&0.86(0.36)&0.61(0.71)&0.61(0.71)&0.98(0.28)&0.33(0.2) \\

 \bottomrule
\end{tabular}
\end{adjustbox}
\caption{An example of TopicalChat-USR and DailyDialog-Grade datasets with evaluation results of selected metrics. We also report the rank score of each metric inside the parentheses, where the rank of the metric score is divided by the total number of evaluated examples.}
\label{tab:case_study}
% \vspace{-0.5cm}
\end{table}

%% file: sections/5_conclusion.tex
\section{Conclusion}
\label{sec:conclusion}
In this work, we propose \textsc{PairEval}, a new comparative evaluation paradigm for assessing the quality of individual responses by considering their relative quality against a few comparison examples.
We encourage moderate-size and open-source LLMs to be specialized for pairwise comparison.
Experiments on multiple evaluation benchmarks demonstrate that \textsc{PairEval} correlates well with human judgments, confirming its effectiveness and validity.
Although effective, \textsc{PairEval} inevitably introduces multiple LLM calls for a single evaluation.
This problem would be amplified as we use a larger set of comparison examples.
Our future work should address an efficient way to find an optimal and small number of comparison examples. 

%% file: sections/A_appendix.tex
\newpage
\section{Appendix}

\subsection{Baseline Details}
\label{secsec:baseline_detail}

We present further implementation details for baselines.

\begin{itemize}
    \item BLEU-2~\citep{papineni-etal-2002-bleu}: We use BLEU-2 score implemented in Natural Language Toolkit~(NLTK) library~\citep{bird2009natural}.

    \item BLEURT~\citep{sellam-etal-2020-bleurt}: We use an \url{Elron/bleurt-tiny-512} in Huggingface Hub.\footnote{\url{https://huggingface.co/Elron/bleurt-tiny-512}}

    \item BERTScore~\citep{zhang2019bertscore}: We use an official implementation\footnote{\url{https://github.com/Tiiiger/bert_score}} by the authors with RoBERTa-large~\citep{liu2019roberta}.

    \item FED~\citep{mehri2020fed}: We use an official implementation\footnote{\url{https://github.com/Shikib/fed}} by the authors with DialoGPT~\citep{zhang2020dialogpt}.

    \item CTC~\citep{ctc}: We use official models and implementation\footnote{\url{https://github.com/tanyuqian/ctc-gen-eval}} by the authors.

    \item USR~\cite{mehri-eskenazi-2020-usr}: We use USR-Retrieval for the overall experiments. For results in TopicalChat-USR and ConvAI2-USR, we cite the results in \citet{mehri-eskenazi-2020-usr}. For the FED dataset, we cite results reported by \cite{yeh-etal-2021-comprehensive}.
    For other meta-evaluation datasets, we use a BERT \citep{devlin-etal-2019-bert} finetuned on a response selection task.
    
    \item FULL~\citep{full}: We use an official implementation\footnote{\url{https://github.com/maximedb/full}} by the authors.

    \item UniEval~\citep{unieval}: We use use official model and implementation\footnote{\url{https://github.com/maszhongming/UniEval}} by the authors.

    \item DEnsity~\citep{park-etal-2023-density}: We use official models and implementation\footnote{\url{https://github.com/ddehun/DEnsity/tree/master}} by the authors.
    
\end{itemize}

\subsection{Additional Qualitative Results}
\label{secsec:additional_scatter}

\begin{figure}[t!]
\centering
\includegraphics[width=0.98\textwidth]{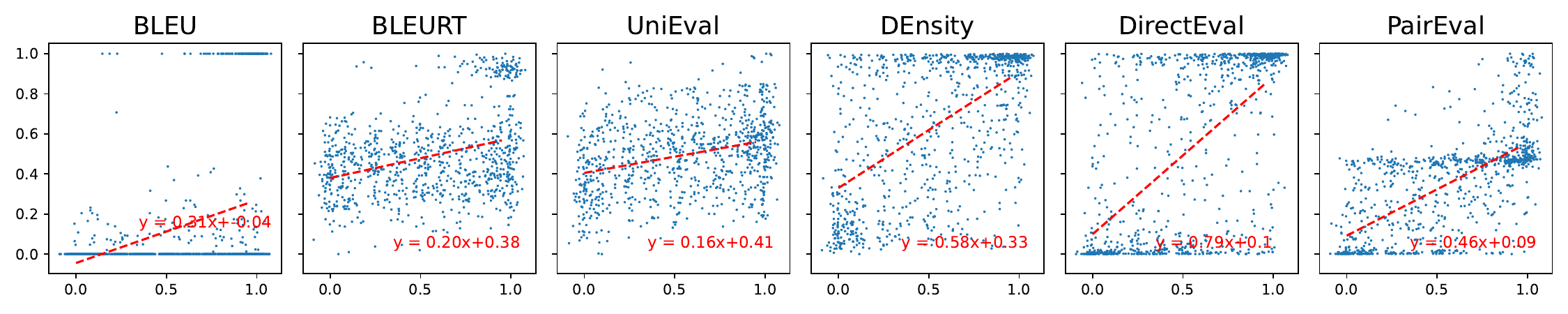}
\caption{
Scatter plots between human judgments and metric scores on the DailyDialog-Zhao dataset. 
The indicators are the same as Fig.~\ref{fig:scatter}.
}
\label{fig:scatter_dd_zhao}
\vspace{-0.3cm}
\end{figure}

\begin{figure}[t!]
\centering
\includegraphics[width=0.98\textwidth]{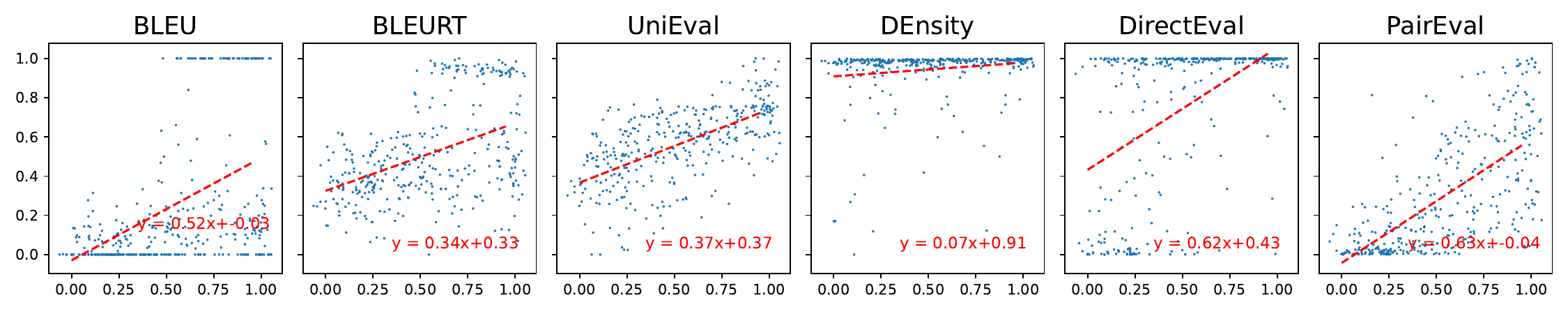}
\caption{
Scatter plots between human judgments and metric scores on the TopicalChat-USR dataset. 
The indicators are the same as Fig.~\ref{fig:scatter}.
}
\label{fig:scatter_topicalchat_usr}
\vspace{-0.3cm}
\end{figure}

\begin{figure}[t!]
\centering
\includegraphics[width=0.9\textwidth]{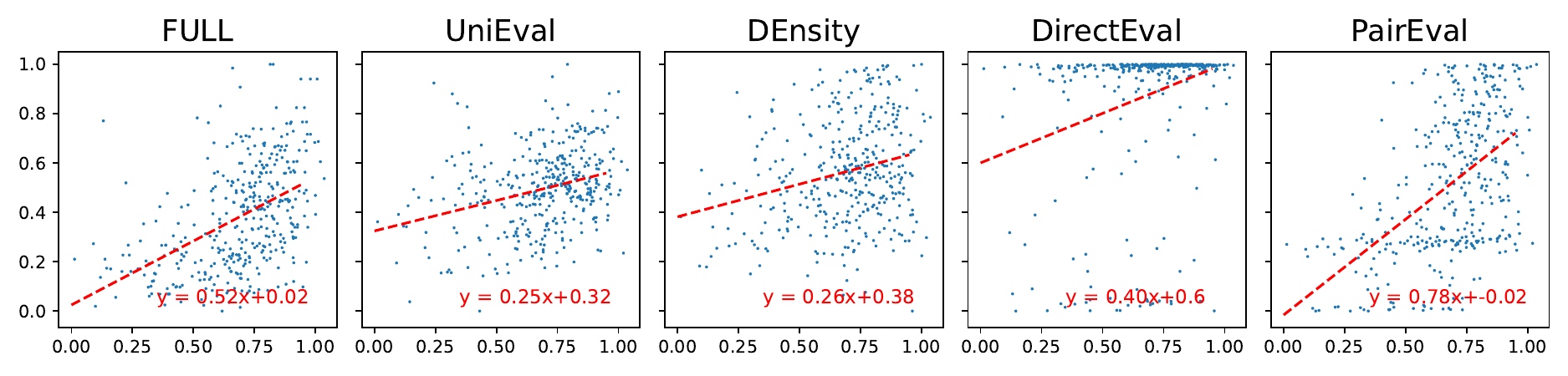}
\caption{
Scatter plots between human judgments and metric scores on the FED dataset. 
The indicators are the same as Fig.~\ref{fig:scatter}.
}
\label{fig:scatter_fed}
\vspace{-0.3cm}
\end{figure}

We present more visualized predictions of selected metrics on other datasets in Fig.~\ref{fig:scatter_dd_zhao}, Fig.~\ref{fig:scatter_topicalchat_usr}, and Fig.~\ref{fig:scatter_fed}. 
From the results in DailyDialog-Zhao dataset (Fig.~\ref{fig:scatter_dd_zhao}), we observe that \textsc{PairEval} often outputs predictions concentrated on a certain range (around 0.5). 
We believe such behaviors can be alleviated through post-hoc calibration techniques like temperature scaling \citep{guo2017calibration} if necessary.

\newpage


%% file: colm2024_conference.bbl
\begin{thebibliography}{55}
\providecommand{\natexlab}[1]{#1}
\providecommand{\url}[1]{\texttt{#1}}
\expandafter\ifx\csname urlstyle\endcsname\relax
  \providecommand{\doi}[1]{doi: #1}\else
  \providecommand{\doi}{doi: \begingroup \urlstyle{rm}\Url}\fi

\bibitem[Bai et~al.(2022)Bai, Jones, Ndousse, Askell, Chen, DasSarma, Drain, Fort, Ganguli, Henighan, et~al.]{bai2022training}
Yuntao Bai, Andy Jones, Kamal Ndousse, Amanda Askell, Anna Chen, Nova DasSarma, Dawn Drain, Stanislav Fort, Deep Ganguli, Tom Henighan, et~al.
\newblock Training a helpful and harmless assistant with reinforcement learning from human feedback.
\newblock \emph{arXiv preprint arXiv:2204.05862}, 2022.

\bibitem[Beutel et~al.(2019)Beutel, Chen, Doshi, Qian, Wei, Wu, Heldt, Zhao, Hong, Chi, et~al.]{beutel2019fairness}
Alex Beutel, Jilin Chen, Tulsee Doshi, Hai Qian, Li~Wei, Yi~Wu, Lukasz Heldt, Zhe Zhao, Lichan Hong, Ed~H Chi, et~al.
\newblock Fairness in recommendation ranking through pairwise comparisons.
\newblock In \emph{Proceedings of the 25th ACM SIGKDD international conference on knowledge discovery \& data mining}, pp.\  2212--2220, 2019.

\bibitem[Bird et~al.(2009)Bird, Klein, and Loper]{bird2009natural}
Steven Bird, Ewan Klein, and Edward Loper.
\newblock \emph{Natural language processing with Python: analyzing text with the natural language toolkit}.
\newblock " O'Reilly Media, Inc.", 2009.

\bibitem[Boubdir et~al.(2023)Boubdir, Kim, Ermis, Hooker, and Fadaee]{boubdir2023elo}
Meriem Boubdir, Edward Kim, Beyza Ermis, Sara Hooker, and Marzieh Fadaee.
\newblock Elo uncovered: Robustness and best practices in language model evaluation.
\newblock \emph{arXiv preprint arXiv:2311.17295}, 2023.

\bibitem[De~Bruyn et~al.(2022)De~Bruyn, Lotfi, Buhmann, and Daelemans]{full}
Maxime De~Bruyn, Ehsan Lotfi, Jeska Buhmann, and Walter Daelemans.
\newblock Open-domain dialog evaluation using follow-ups likelihood.
\newblock In Nicoletta Calzolari, Chu-Ren Huang, Hansaem Kim, James Pustejovsky, Leo Wanner, Key-Sun Choi, Pum-Mo Ryu, Hsin-Hsi Chen, Lucia Donatelli, Heng Ji, Sadao Kurohashi, Patrizia Paggio, Nianwen Xue, Seokhwan Kim, Younggyun Hahm, Zhong He, Tony~Kyungil Lee, Enrico Santus, Francis Bond, and Seung-Hoon Na (eds.), \emph{Proceedings of the 29th International Conference on Computational Linguistics}, pp.\  496--504, Gyeongju, Republic of Korea, October 2022. International Committee on Computational Linguistics.
\newblock URL \url{https://aclanthology.org/2022.coling-1.40}.

\bibitem[Deng et~al.(2021)Deng, Tan, Liu, Xing, and Hu]{ctc}
Mingkai Deng, Bowen Tan, Zhengzhong Liu, Eric Xing, and Zhiting Hu.
\newblock Compression, transduction, and creation: A unified framework for evaluating natural language generation.
\newblock In Marie-Francine Moens, Xuanjing Huang, Lucia Specia, and Scott Wen-tau Yih (eds.), \emph{Proceedings of the 2021 Conference on Empirical Methods in Natural Language Processing}, pp.\  7580--7605, Online and Punta Cana, Dominican Republic, November 2021. Association for Computational Linguistics.
\newblock \doi{10.18653/v1/2021.emnlp-main.599}.
\newblock URL \url{https://aclanthology.org/2021.emnlp-main.599}.

\bibitem[Dettmers et~al.(2024)Dettmers, Pagnoni, Holtzman, and Zettlemoyer]{dettmers2024qlora}
Tim Dettmers, Artidoro Pagnoni, Ari Holtzman, and Luke Zettlemoyer.
\newblock Qlora: Efficient finetuning of quantized llms.
\newblock \emph{Advances in Neural Information Processing Systems}, 36, 2024.

\bibitem[Devlin et~al.(2019)Devlin, Chang, Lee, and Toutanova]{devlin-etal-2019-bert}
Jacob Devlin, Ming-Wei Chang, Kenton Lee, and Kristina Toutanova.
\newblock {BERT}: Pre-training of deep bidirectional transformers for language understanding.
\newblock In \emph{Proceedings of the 2019 Conference of the North {A}merican Chapter of the Association for Computational Linguistics: Human Language Technologies, Volume 1 (Long and Short Papers)}, pp.\  4171--4186, Minneapolis, Minnesota, June 2019. Association for Computational Linguistics.
\newblock \doi{10.18653/v1/N19-1423}.
\newblock URL \url{https://aclanthology.org/N19-1423}.

\bibitem[Dinan et~al.(2020)Dinan, Logacheva, Malykh, Miller, Shuster, Urbanek, Kiela, Szlam, Serban, Lowe, et~al.]{dinan2020second}
Emily Dinan, Varvara Logacheva, Valentin Malykh, Alexander Miller, Kurt Shuster, Jack Urbanek, Douwe Kiela, Arthur Szlam, Iulian Serban, Ryan Lowe, et~al.
\newblock The second conversational intelligence challenge (convai2).
\newblock In \emph{The NeurIPS'18 Competition}, pp.\  187--208. Springer, 2020.

\bibitem[F{\"u}rnkranz \& H{\"u}llermeier(2003)F{\"u}rnkranz and H{\"u}llermeier]{furnkranz2003pairwise}
Johannes F{\"u}rnkranz and Eyke H{\"u}llermeier.
\newblock Pairwise preference learning and ranking.
\newblock In \emph{European conference on machine learning}, pp.\  145--156. Springer, 2003.

\bibitem[Ghazarian et~al.(2019)Ghazarian, Wei, Galstyan, and Peng]{ghazarian-etal-2019-better}
Sarik Ghazarian, Johnny Wei, Aram Galstyan, and Nanyun Peng.
\newblock Better automatic evaluation of open-domain dialogue systems with contextualized embeddings.
\newblock In \emph{Proceedings of the Workshop on Methods for Optimizing and Evaluating Neural Language Generation}, pp.\  82--89, Minneapolis, Minnesota, June 2019. Association for Computational Linguistics.
\newblock \doi{10.18653/v1/W19-2310}.
\newblock URL \url{https://aclanthology.org/W19-2310}.

\bibitem[Gopalakrishnan et~al.(2019)Gopalakrishnan, Hedayatnia, Chen, Gottardi, Kwatra, Venkatesh, Gabriel, and Hakkani-Tür]{gopalakrishnan2019topical}
Karthik Gopalakrishnan, Behnam Hedayatnia, Qinlang Chen, Anna Gottardi, Sanjeev Kwatra, Anu Venkatesh, Raefer Gabriel, and Dilek Hakkani-Tür.
\newblock {Topical-Chat: Towards Knowledge-Grounded Open-Domain Conversations}.
\newblock In \emph{Proc. Interspeech 2019}, pp.\  1891--1895, 2019.
\newblock \doi{10.21437/Interspeech.2019-3079}.
\newblock URL \url{http://dx.doi.org/10.21437/Interspeech.2019-3079}.

\bibitem[Guo et~al.(2017)Guo, Pleiss, Sun, and Weinberger]{guo2017calibration}
Chuan Guo, Geoff Pleiss, Yu~Sun, and Kilian~Q Weinberger.
\newblock On calibration of modern neural networks.
\newblock In \emph{International conference on machine learning}, pp.\  1321--1330. PMLR, 2017.

\bibitem[Gupta et~al.(2021)Gupta, Tsvetkov, and Bigham]{gupta-etal-2021-synthesizing}
Prakhar Gupta, Yulia Tsvetkov, and Jeffrey Bigham.
\newblock Synthesizing adversarial negative responses for robust response ranking and evaluation.
\newblock In \emph{Findings of the Association for Computational Linguistics: ACL-IJCNLP 2021}, pp.\  3867--3883, Online, August 2021. Association for Computational Linguistics.
\newblock \doi{10.18653/v1/2021.findings-acl.338}.
\newblock URL \url{https://aclanthology.org/2021.findings-acl.338}.

\bibitem[Gupta et~al.(2022)Gupta, Jiao, Yeh, Mehri, Eskenazi, and Bigham]{gupta2022instructdial}
Prakhar Gupta, Cathy Jiao, Yi-Ting Yeh, Shikib Mehri, Maxine Eskenazi, and Jeffrey~P Bigham.
\newblock Instructdial: Improving zero and few-shot generalization in dialogue through instruction tuning.
\newblock In \emph{Proceedings of the 2022 Conference on Empirical Methods in Natural Language Processing}, pp.\  505--525, 2022.

\bibitem[Hu et~al.(2021)Hu, Wallis, Allen-Zhu, Li, Wang, Wang, Chen, et~al.]{hu2021lora}
Edward~J Hu, Phillip Wallis, Zeyuan Allen-Zhu, Yuanzhi Li, Shean Wang, Lu~Wang, Weizhu Chen, et~al.
\newblock Lora: Low-rank adaptation of large language models.
\newblock In \emph{International Conference on Learning Representations}, 2021.

\bibitem[Huang et~al.(2020)Huang, Ye, Qin, Lin, and Liang]{huang2020grade}
Lishan Huang, Zheng Ye, Jinghui Qin, Liang Lin, and Xiaodan Liang.
\newblock {GRADE}: Automatic graph-enhanced coherence metric for evaluating open-domain dialogue systems.
\newblock In \emph{Proceedings of the 2020 Conference on Empirical Methods in Natural Language Processing (EMNLP)}, pp.\  9230--9240, Online, November 2020. Association for Computational Linguistics.
\newblock \doi{10.18653/v1/2020.emnlp-main.742}.
\newblock URL \url{https://aclanthology.org/2020.emnlp-main.742}.

\bibitem[Kendall(1955)]{kendall1955rank}
Maurice~G Kendall.
\newblock Rank correlation methods. new york: Hafner, 1955.
\newblock \emph{Manuscript received 3/30}, 65, 1955.

\bibitem[Khalid \& Lee(2022)Khalid and Lee]{khalid-lee-2022-explaining}
Baber Khalid and Sungjin Lee.
\newblock Explaining dialogue evaluation metrics using adversarial behavioral analysis.
\newblock In \emph{Proceedings of the 2022 Conference of the North American Chapter of the Association for Computational Linguistics: Human Language Technologies}, pp.\  5871--5883, Seattle, United States, July 2022. Association for Computational Linguistics.
\newblock \doi{10.18653/v1/2022.naacl-main.430}.
\newblock URL \url{https://aclanthology.org/2022.naacl-main.430}.

\bibitem[Lee et~al.(2022)Lee, Park, Choi, and Choo]{lee-etal-2022-pneg}
Nyoungwoo Lee, ChaeHun Park, Ho-Jin Choi, and Jaegul Choo.
\newblock Pneg: Prompt-based negative response generation for dialogue response selection task.
\newblock In \emph{Proceedings of the 2022 Conference on Empirical Methods in Natural Language Processing}, pp.\  10692--10703, Abu Dhabi, United Arab Emirates, December 2022. Association for Computational Linguistics.
\newblock URL \url{https://aclanthology.org/2022.emnlp-main.733}.

\bibitem[Li et~al.(2017)Li, Su, Shen, Li, Cao, and Niu]{li-etal-2017-dailydialog}
Yanran Li, Hui Su, Xiaoyu Shen, Wenjie Li, Ziqiang Cao, and Shuzi Niu.
\newblock {D}aily{D}ialog: A manually labelled multi-turn dialogue dataset.
\newblock In \emph{Proceedings of the Eighth International Joint Conference on Natural Language Processing (Volume 1: Long Papers)}, pp.\  986--995, Taipei, Taiwan, November 2017. Asian Federation of Natural Language Processing.
\newblock URL \url{https://aclanthology.org/I17-1099}.

\bibitem[Liang et~al.(2023)Liang, Bommasani, Lee, Tsipras, Soylu, Yasunaga, Zhang, Narayanan, Wu, Kumar, et~al.]{liang2023holistic}
Percy Liang, Rishi Bommasani, Tony Lee, Dimitris Tsipras, Dilara Soylu, Michihiro Yasunaga, Yian Zhang, Deepak Narayanan, Yuhuai Wu, Ananya Kumar, et~al.
\newblock Holistic evaluation of language models.
\newblock \emph{Transactions on Machine Learning Research}, 2023.

\bibitem[Lin(2004)]{lin-2004-rouge}
Chin-Yew Lin.
\newblock {ROUGE}: A package for automatic evaluation of summaries.
\newblock In \emph{Text Summarization Branches Out}, pp.\  74--81, Barcelona, Spain, July 2004. Association for Computational Linguistics.
\newblock URL \url{https://aclanthology.org/W04-1013}.

\bibitem[Lin \& Chen(2023)Lin and Chen]{llmeval}
Yen-Ting Lin and Yun-Nung Chen.
\newblock {LLM}-eval: Unified multi-dimensional automatic evaluation for open-domain conversations with large language models.
\newblock In Yun-Nung Chen and Abhinav Rastogi (eds.), \emph{Proceedings of the 5th Workshop on NLP for Conversational AI (NLP4ConvAI 2023)}, pp.\  47--58, Toronto, Canada, July 2023. Association for Computational Linguistics.
\newblock \doi{10.18653/v1/2023.nlp4convai-1.5}.
\newblock URL \url{https://aclanthology.org/2023.nlp4convai-1.5}.

\bibitem[Liu et~al.(2022)Liu, Swayamdipta, Smith, and Choi]{liu-etal-2022-wanli}
Alisa Liu, Swabha Swayamdipta, Noah~A. Smith, and Yejin Choi.
\newblock {WANLI}: Worker and {AI} collaboration for natural language inference dataset creation.
\newblock In Yoav Goldberg, Zornitsa Kozareva, and Yue Zhang (eds.), \emph{Findings of the Association for Computational Linguistics: EMNLP 2022}, pp.\  6826--6847, Abu Dhabi, United Arab Emirates, December 2022. Association for Computational Linguistics.
\newblock \doi{10.18653/v1/2022.findings-emnlp.508}.
\newblock URL \url{https://aclanthology.org/2022.findings-emnlp.508}.

\bibitem[Liu et~al.(2016)Liu, Lowe, Serban, Noseworthy, Charlin, and Pineau]{liu-etal-2016-evaluate}
Chia-Wei Liu, Ryan Lowe, Iulian Serban, Mike Noseworthy, Laurent Charlin, and Joelle Pineau.
\newblock How {NOT} to evaluate your dialogue system: An empirical study of unsupervised evaluation metrics for dialogue response generation.
\newblock In \emph{Proceedings of the 2016 Conference on Empirical Methods in Natural Language Processing}, pp.\  2122--2132, Austin, Texas, November 2016. Association for Computational Linguistics.
\newblock \doi{10.18653/v1/D16-1230}.
\newblock URL \url{https://aclanthology.org/D16-1230}.

\bibitem[Liu et~al.(2019)Liu, Ott, Goyal, Du, Joshi, Chen, Levy, Lewis, Zettlemoyer, and Stoyanov]{liu2019roberta}
Yinhan Liu, Myle Ott, Naman Goyal, Jingfei Du, Mandar Joshi, Danqi Chen, Omer Levy, Mike Lewis, Luke Zettlemoyer, and Veselin Stoyanov.
\newblock Roberta: A robustly optimized bert pretraining approach.
\newblock \emph{arXiv preprint arXiv:1907.11692}, 2019.

\bibitem[Liusie et~al.(2024)Liusie, Manakul, and Gales]{liusie-comparative}
Adian Liusie, Potsawee Manakul, and Mark Gales.
\newblock {LLM} comparative assessment: Zero-shot {NLG} evaluation through pairwise comparisons using large language models.
\newblock In Yvette Graham and Matthew Purver (eds.), \emph{Proceedings of the 18th Conference of the European Chapter of the Association for Computational Linguistics (Volume 1: Long Papers)}, pp.\  139--151, St. Julian{'}s, Malta, March 2024. Association for Computational Linguistics.
\newblock URL \url{https://aclanthology.org/2024.eacl-long.8}.

\bibitem[Loshchilov \& Hutter(2018)Loshchilov and Hutter]{adamw}
Ilya Loshchilov and Frank Hutter.
\newblock Decoupled weight decay regularization.
\newblock In \emph{International Conference on Learning Representations}, 2018.

\bibitem[Lowe et~al.(2017)Lowe, Noseworthy, Serban, Angelard-Gontier, Bengio, and Pineau]{lowe-etal-2017-adem}
Ryan Lowe, Michael Noseworthy, Iulian~Vlad Serban, Nicolas Angelard-Gontier, Yoshua Bengio, and Joelle Pineau.
\newblock Towards an automatic {T}uring test: Learning to evaluate dialogue responses.
\newblock In \emph{Proceedings of the 55th Annual Meeting of the Association for Computational Linguistics (Volume 1: Long Papers)}, pp.\  1116--1126, Vancouver, Canada, July 2017. Association for Computational Linguistics.
\newblock \doi{10.18653/v1/P17-1103}.
\newblock URL \url{https://aclanthology.org/P17-1103}.

\bibitem[Mehri \& Eskenazi(2020{\natexlab{a}})Mehri and Eskenazi]{mehri-eskenazi-2020-usr}
Shikib Mehri and Maxine Eskenazi.
\newblock {USR}: An unsupervised and reference free evaluation metric for dialog generation.
\newblock In \emph{Proceedings of the 58th Annual Meeting of the Association for Computational Linguistics}, pp.\  681--707, Online, July 2020{\natexlab{a}}. Association for Computational Linguistics.
\newblock \doi{10.18653/v1/2020.acl-main.64}.
\newblock URL \url{https://aclanthology.org/2020.acl-main.64}.

\bibitem[Mehri \& Eskenazi(2020{\natexlab{b}})Mehri and Eskenazi]{mehri2020fed}
Shikib Mehri and Maxine Eskenazi.
\newblock Unsupervised evaluation of interactive dialog with dialogpt.
\newblock In \emph{Proceedings of the 21th Annual Meeting of the Special Interest Group on Discourse and Dialogue}, pp.\  225--235, 2020{\natexlab{b}}.

\bibitem[Ouyang et~al.(2022)Ouyang, Wu, Jiang, Almeida, Wainwright, Mishkin, Zhang, Agarwal, Slama, Ray, et~al.]{ouyang2022training}
Long Ouyang, Jeffrey Wu, Xu~Jiang, Diogo Almeida, Carroll Wainwright, Pamela Mishkin, Chong Zhang, Sandhini Agarwal, Katarina Slama, Alex Ray, et~al.
\newblock Training language models to follow instructions with human feedback.
\newblock \emph{Advances in neural information processing systems}, 35:\penalty0 27730--27744, 2022.

\bibitem[Papineni et~al.(2002)Papineni, Roukos, Ward, and Zhu]{papineni-etal-2002-bleu}
Kishore Papineni, Salim Roukos, Todd Ward, and Wei-Jing Zhu.
\newblock {B}leu: a method for automatic evaluation of machine translation.
\newblock In \emph{Proceedings of the 40th Annual Meeting of the Association for Computational Linguistics}, pp.\  311--318, Philadelphia, Pennsylvania, USA, July 2002. Association for Computational Linguistics.
\newblock \doi{10.3115/1073083.1073135}.
\newblock URL \url{https://aclanthology.org/P02-1040}.

\bibitem[Park et~al.(2021)Park, Jang, Yang, and Park]{park-etal-2021-generating}
ChaeHun Park, Eugene Jang, Wonsuk Yang, and Jong Park.
\newblock Generating negative samples by manipulating golden responses for unsupervised learning of a response evaluation model.
\newblock In \emph{Proceedings of the 2021 Conference of the North American Chapter of the Association for Computational Linguistics: Human Language Technologies}, pp.\  1525--1534, Online, June 2021. Association for Computational Linguistics.
\newblock \doi{10.18653/v1/2021.naacl-main.120}.
\newblock URL \url{https://aclanthology.org/2021.naacl-main.120}.

\bibitem[Park et~al.(2023)Park, Lee, Rim, and Choo]{park-etal-2023-density}
ChaeHun Park, Seungil Lee, Daniel Rim, and Jaegul Choo.
\newblock {DE}nsity: Open-domain dialogue evaluation metric using density estimation.
\newblock In Anna Rogers, Jordan Boyd-Graber, and Naoaki Okazaki (eds.), \emph{Findings of the Association for Computational Linguistics: ACL 2023}, pp.\  14222--14236, Toronto, Canada, July 2023. Association for Computational Linguistics.
\newblock \doi{10.18653/v1/2023.findings-acl.896}.
\newblock URL \url{https://aclanthology.org/2023.findings-acl.896}.

\bibitem[Pillutla et~al.(2021)Pillutla, Swayamdipta, Zellers, Thickstun, Welleck, Choi, and Harchaoui]{pillutla2021mauve}
Krishna Pillutla, Swabha Swayamdipta, Rowan Zellers, John Thickstun, Sean Welleck, Yejin Choi, and Zaid Harchaoui.
\newblock Mauve: Measuring the gap between neural text and human text using divergence frontiers.
\newblock \emph{Advances in Neural Information Processing Systems}, 34:\penalty0 4816--4828, 2021.

\bibitem[Qin et~al.(2023)Qin, Jagerman, Hui, Zhuang, Wu, Shen, Liu, Liu, Metzler, Wang, et~al.]{qin2023large}
Zhen Qin, Rolf Jagerman, Kai Hui, Honglei Zhuang, Junru Wu, Jiaming Shen, Tianqi Liu, Jialu Liu, Donald Metzler, Xuanhui Wang, et~al.
\newblock Large language models are effective text rankers with pairwise ranking prompting.
\newblock \emph{arXiv preprint arXiv:2306.17563}, 2023.

\bibitem[Sai et~al.(2020)Sai, Mohankumar, Arora, and Khapra]{sai-etal-2020-deb}
Ananya~B. Sai, Akash~Kumar Mohankumar, Siddhartha Arora, and Mitesh~M. Khapra.
\newblock Improving dialog evaluation with a multi-reference adversarial dataset and large scale pretraining.
\newblock \emph{Transactions of the Association for Computational Linguistics}, 8:\penalty0 810--827, 2020.
\newblock \doi{10.1162/tacl_a_00347}.
\newblock URL \url{https://aclanthology.org/2020.tacl-1.52}.

\bibitem[Sato et~al.(2020)Sato, Akama, Ouchi, Suzuki, and Inui]{sato-etal-2020-evaluating}
Shiki Sato, Reina Akama, Hiroki Ouchi, Jun Suzuki, and Kentaro Inui.
\newblock Evaluating dialogue generation systems via response selection.
\newblock In Dan Jurafsky, Joyce Chai, Natalie Schluter, and Joel Tetreault (eds.), \emph{Proceedings of the 58th Annual Meeting of the Association for Computational Linguistics}, pp.\  593--599, Online, July 2020. Association for Computational Linguistics.
\newblock \doi{10.18653/v1/2020.acl-main.55}.
\newblock URL \url{https://aclanthology.org/2020.acl-main.55}.

\bibitem[Sellam et~al.(2020)Sellam, Das, and Parikh]{sellam-etal-2020-bleurt}
Thibault Sellam, Dipanjan Das, and Ankur Parikh.
\newblock {BLEURT}: Learning robust metrics for text generation.
\newblock In \emph{Proceedings of the 58th Annual Meeting of the Association for Computational Linguistics}, pp.\  7881--7892, Online, July 2020. Association for Computational Linguistics.
\newblock \doi{10.18653/v1/2020.acl-main.704}.
\newblock URL \url{https://aclanthology.org/2020.acl-main.704}.

\bibitem[Tao et~al.(2018)Tao, Mou, Zhao, and Yan]{tao2018ruber}
Chongyang Tao, Lili Mou, Dongyan Zhao, and Rui Yan.
\newblock Ruber: An unsupervised method for automatic evaluation of open-domain dialog systems.
\newblock In \emph{Thirty-Second AAAI Conference on Artificial Intelligence}, 2018.

\bibitem[Touvron et~al.(2023)Touvron, Martin, Stone, Albert, Almahairi, Babaei, Bashlykov, Batra, Bhargava, Bhosale, et~al.]{llama-2-paper}
Hugo Touvron, Louis Martin, Kevin Stone, Peter Albert, Amjad Almahairi, Yasmine Babaei, Nikolay Bashlykov, Soumya Batra, Prajjwal Bhargava, Shruti Bhosale, et~al.
\newblock Llama 2: Open foundation and fine-tuned chat models.
\newblock \emph{arXiv preprint arXiv:2307.09288}, 2023.

\bibitem[Wang et~al.(2023)Wang, Li, Chen, Zhu, Lin, Cao, Liu, Liu, and Sui]{LLM_unfair_eval}
Peiyi Wang, Lei Li, Liang Chen, Dawei Zhu, Binghuai Lin, Yunbo Cao, Qi~Liu, Tianyu Liu, and Zhifang Sui.
\newblock Large language models are not fair evaluators.
\newblock \emph{arXiv preprint arXiv:2305.17926}, 2023.

\bibitem[Xu et~al.(2020)Xu, Wei, Xia, Lan, Yin, Cheng, and Wen]{xu2020reinforcement}
Jun Xu, Zeng Wei, Long Xia, Yanyan Lan, Dawei Yin, Xueqi Cheng, and Ji-Rong Wen.
\newblock Reinforcement learning to rank with pairwise policy gradient.
\newblock In \emph{Proceedings of the 43rd International ACM SIGIR Conference on Research and Development in Information Retrieval}, pp.\  509--518, 2020.

\bibitem[Yeh et~al.(2021)Yeh, Eskenazi, and Mehri]{yeh-etal-2021-comprehensive}
Yi-Ting Yeh, Maxine Eskenazi, and Shikib Mehri.
\newblock A comprehensive assessment of dialog evaluation metrics.
\newblock In Wei Wei, Bo~Dai, Tuo Zhao, Lihong Li, Diyi Yang, Yun-Nung Chen, Y-Lan Boureau, Asli Celikyilmaz, Alborz Geramifard, Aman Ahuja, and Haoming Jiang (eds.), \emph{The First Workshop on Evaluations and Assessments of Neural Conversation Systems}, pp.\  15--33, Online, November 2021. Association for Computational Linguistics.
\newblock \doi{10.18653/v1/2021.eancs-1.3}.
\newblock URL \url{https://aclanthology.org/2021.eancs-1.3}.

\bibitem[Yoo et~al.(2021)Yoo, Park, Kang, Lee, and Park]{yoo-etal-2021-gpt3mix-leveraging}
Kang~Min Yoo, Dongju Park, Jaewook Kang, Sang-Woo Lee, and Woomyoung Park.
\newblock {GPT}3{M}ix: Leveraging large-scale language models for text augmentation.
\newblock In Marie-Francine Moens, Xuanjing Huang, Lucia Specia, and Scott Wen-tau Yih (eds.), \emph{Findings of the Association for Computational Linguistics: EMNLP 2021}, pp.\  2225--2239, Punta Cana, Dominican Republic, November 2021. Association for Computational Linguistics.
\newblock \doi{10.18653/v1/2021.findings-emnlp.192}.
\newblock URL \url{https://aclanthology.org/2021.findings-emnlp.192}.

\bibitem[Zar(2005)]{zar2005spearman}
Jerrold~H Zar.
\newblock Spearman rank correlation.
\newblock \emph{Encyclopedia of Biostatistics}, 7, 2005.

\bibitem[Zhang et~al.(2021)Zhang, Chen, D{'}Haro, Zhang, Friedrichs, Lee, and Li]{zhang-etal-2021-dynaeval}
Chen Zhang, Yiming Chen, Luis~Fernando D{'}Haro, Yan Zhang, Thomas Friedrichs, Grandee Lee, and Haizhou Li.
\newblock {D}yna{E}val: Unifying turn and dialogue level evaluation.
\newblock In \emph{Proceedings of the 59th Annual Meeting of the Association for Computational Linguistics and the 11th International Joint Conference on Natural Language Processing (Volume 1: Long Papers)}, pp.\  5676--5689, Online, August 2021. Association for Computational Linguistics.
\newblock \doi{10.18653/v1/2021.acl-long.441}.
\newblock URL \url{https://aclanthology.org/2021.acl-long.441}.

\bibitem[Zhang et~al.(2023)Zhang, D'Haro, Chen, Zhang, and Li]{zhang2023comprehensive}
Chen Zhang, Luis~Fernando D'Haro, Yiming Chen, Malu Zhang, and Haizhou Li.
\newblock A comprehensive analysis of the effectiveness of large language models as automatic dialogue evaluators.
\newblock \emph{arXiv preprint arXiv:2312.15407}, 2023.

\bibitem[Zhang et~al.(2019)Zhang, Kishore, Wu, Weinberger, and Artzi]{zhang2019bertscore}
Tianyi Zhang, Varsha Kishore, Felix Wu, Kilian~Q Weinberger, and Yoav Artzi.
\newblock Bertscore: Evaluating text generation with bert.
\newblock In \emph{International Conference on Learning Representations}, 2019.

\bibitem[Zhang et~al.(2020)Zhang, Sun, Galley, Chen, Brockett, Gao, Gao, Liu, and Dolan]{zhang2020dialogpt}
Yizhe Zhang, Siqi Sun, Michel Galley, Yen-Chun Chen, Chris Brockett, Xiang Gao, Jianfeng Gao, Jingjing Liu, and William~B Dolan.
\newblock Dialogpt: Large-scale generative pre-training for conversational response generation.
\newblock In \emph{Proceedings of the 58th Annual Meeting of the Association for Computational Linguistics: System Demonstrations}, pp.\  270--278, 2020.

\bibitem[Zhao et~al.(2020)Zhao, Lala, and Kawahara]{zhao2020designing}
Tianyu Zhao, Divesh Lala, and Tatsuya Kawahara.
\newblock Designing precise and robust dialogue response evaluators.
\newblock In \emph{Proceedings of the 58th Annual Meeting of the Association for Computational Linguistics}, pp.\  26--33, Online, July 2020. Association for Computational Linguistics.
\newblock \doi{10.18653/v1/2020.acl-main.4}.
\newblock URL \url{https://aclanthology.org/2020.acl-main.4}.

\bibitem[Zheng et~al.(2024)Zheng, Chiang, Sheng, Zhuang, Wu, Zhuang, Lin, Li, Li, Xing, et~al.]{zheng2024judging}
Lianmin Zheng, Wei-Lin Chiang, Ying Sheng, Siyuan Zhuang, Zhanghao Wu, Yonghao Zhuang, Zi~Lin, Zhuohan Li, Dacheng Li, Eric Xing, et~al.
\newblock Judging llm-as-a-judge with mt-bench and chatbot arena.
\newblock \emph{Advances in Neural Information Processing Systems}, 36, 2024.

\bibitem[Zhong et~al.(2022)Zhong, Liu, Yin, Mao, Jiao, Liu, Zhu, Ji, and Han]{unieval}
Ming Zhong, Yang Liu, Da~Yin, Yuning Mao, Yizhu Jiao, Pengfei Liu, Chenguang Zhu, Heng Ji, and Jiawei Han.
\newblock Towards a unified multi-dimensional evaluator for text generation.
\newblock In Yoav Goldberg, Zornitsa Kozareva, and Yue Zhang (eds.), \emph{Proceedings of the 2022 Conference on Empirical Methods in Natural Language Processing}, pp.\  2023--2038, Abu Dhabi, United Arab Emirates, December 2022. Association for Computational Linguistics.
\newblock \doi{10.18653/v1/2022.emnlp-main.131}.
\newblock URL \url{https://aclanthology.org/2022.emnlp-main.131}.

\end{thebibliography}
